\documentclass[runningheads]{llncs}
\usepackage{graphicx}
\usepackage{amsmath,amssymb} 
\usepackage{color}

\usepackage[dvipsnames]{xcolor}
\definecolor{b}{RGB}{108,123,252}
\definecolor{g}{RGB}{111, 178, 113}
\usepackage{algorithm} 
\usepackage{algpseudocode} 
\usepackage{subcaption}
\usepackage{wrapfig}
\usepackage[outercaption]{sidecap}
\usepackage{siunitx}
\usepackage{color}
\usepackage{booktabs}
\usepackage{multirow}
\usepackage{caption}
\usepackage{verbatim} 

\begin{document}
\pagestyle{headings}
\mainmatter

\def\ACCV20SubNumber{10}  

\title{RealSmileNet: A Deep End-To-End Network for Spontaneous and Posed Smile Recognition} 
\titlerunning{RealSmileNet}
%
\author{Yan Yang\inst{1}\orcidID{0000-0002-6246-1748} \and
Md Zakir Hossain\inst{1,2}\orcidID{0000-0003-1892-831X} \and
Tom Gedeon\inst{1}\orcidID{0000-0001-8356-4909}\and Shafin Rahman  \inst{1,3,4} \orcidID{0000-0001-7169-0318}}
\authorrunning{Y. Yang et al.}
%
\institute{The Australian National University, Canberra ACT 0200, AU \and The University of Canberra, Bruce ACT 2617, AU \and North South University, Dhaka, Bangladesh \and Data61-CSIRO, Canberra ACT 0200, AU\\
\email{\{u6169130, zakir.hossain, tom.gedeon\}@anu.edu.au, shafin.rahman@northsouth.edu}}

\maketitle

\begin{abstract}

Smiles play a vital role in the understanding of social interactions within different communities, and reveal the physical state of mind of people in both real and deceptive ways. Several methods have been proposed to recognize spontaneous and posed smiles. All follow a feature-engineering based pipeline requiring costly pre-processing steps such as manual annotation of face landmarks, tracking, segmentation of smile phases, and hand-crafted features. The resulting computation is expensive, and strongly dependent on pre-processing steps. We investigate an end-to-end deep learning model to address these problems, the first end-to-end model for spontaneous and posed smile recognition. Our fully automated model is fast and learns the feature extraction processes by training a series of convolution and ConvLSTM layer from scratch. Our experiments on four datasets demonstrate the robustness and generalization of the proposed model by achieving state-of-the-art performances.

\end{abstract}

\section{Introduction}
Facial expression recognition is a process of identifying human emotion from videos, audios, and even the texts. Understanding facial expressions is essential for various forms of communication, such as the interaction between humans and machines. Also, the development of facial expression recognition contributes to the area of market research, health care, video game testings, {and so on} \cite{survey}. Meanwhile, people {tend} to hide their natural expression in different environments. Recognising spontaneous and posed facial expressions {are} necessary for social interaction analysis \cite{SM:2} because it can be deceptive and convey diverse meanings. The \textit{smile} is the most common and easily expressible facial display, but still very hard to recognise. Because of the recurrence and cultural reasons, the study of cognitive and computer sciences broadly investigates the recognition of spontaneous (genuine/real/felt) and posed (fake/false/deliberate) smiles \cite{SM:2,SM:3,SM:4,SM:5,SM:6,cohn,syssmile,CLBPTOP,zk1,zk3,zk4,zk5,zk6}.

\begin{figure}[!t]
    \centering
   \includegraphics[width=1\textwidth,trim={0cm 1.2cm .6cm 0cm},clip]{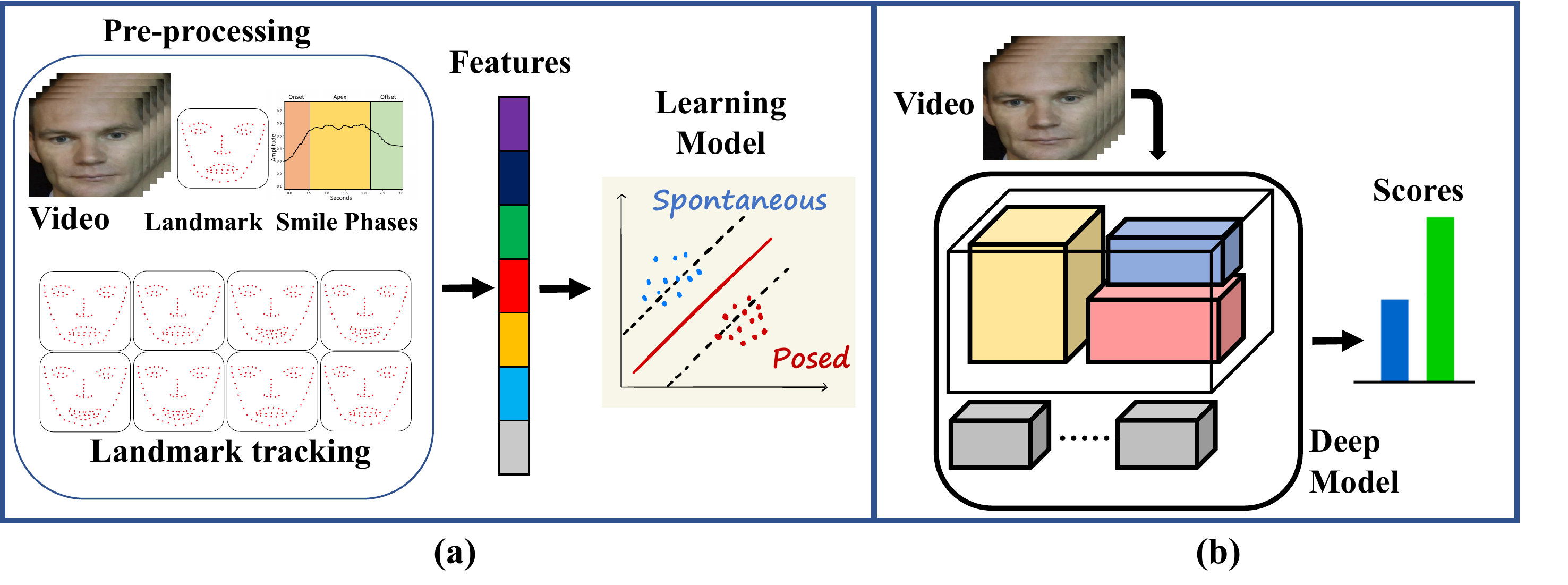}
   \caption{Overview of different spontaneous smile recognition models. \textbf{\textit{(left)} }Given a video as input, previous approaches \cite{SM:2,SM:4,SM:5,SM:6,cohn,CLBPTOP} perform several manual or semi-automatic prepossessing steps like facial landmark detection \cite{SM:2,SM:4,SM:5,SM:6,cohn}, tracking \cite{SM:2,SM:4,SM:5,SM:6,cohn}, smile phases segmentation \cite{SM:2,SM:5}, {and so on}. across frames to calculate hand-engineered feature vectors (D-marker \cite{SM:2,SM:4,DMarker}, HoG \cite{SM:5,CLBPTOP,accv2016}), then feed the features to a learning model (SVM) for classification. The costly intermediate steps significantly increase the computation and limit the fully automatic process. \textbf{\textit{(right)}} Our proposed end-to-end architecture takes video frames as input and recognizes spontaneous and posed smile by a simple forward pass.} 
\label{fig:motivation}
\end{figure}

Previous efforts on recognizing spontaneous and posed smiles mostly follow a feature-based approach where machine learning (ML) models perform a binary classification based on the extracted visual features from a smile video \cite{SM:2,SM:3,SM:4,SM:5,SM:6,cohn,syssmile,CLBPTOP,zk1,zk3,zk4,zk5,accv2016,DMarker}. We identify several limitations of such approaches. \textit{(a) Manual annotation}: Many methods require manual annotation of facial landmarks for the first frame of a video \cite{SM:2,SM:4,SM:5,SM:6,cohn,DMarker}. It limits the automation of the recognition process. \textit{(b) Landmark tracking}: Methods need to track face landmarks throughout the video \cite{SM:2,SM:4,SM:5,SM:6,cohn,DMarker}. It is a computationally expensive process, and the performance of the recognition broadly depends on it. \textit{(c) Segmentation of temporal phases}: Some methods extract features from temporal stages of a smile (i.e., onset, apex, and offset) separately \cite{SM:2,SM:5}. Automatic segmentation of a smile can be erroneous because, in many smile videos, these phases are not apparent, and methods need to assign zero values in the feature list to satisfy the constant length of the feature set. \textit{(d) Limiting the maximum length of a smile}: {Most traditional machine learning methods cannot handle the dynamic length of time series data.} Traditional methods need to represent each smile by a fixed length. It decreases the robustness of the system because, in a real application, a smile video may come with variable length. \textit{(e) Hand-engineered features}: Methods depend on hand-crafted features like D-marker \cite{SM:2,SM:4,DMarker}, Histogram of Oriented Gradients (HoG) \cite{SM:5,CLBPTOP,accv2016}, Local Binary Pattern (LBP) like feature on region of interest \cite{SM:5,accv2016}. The selection of such features sometimes requires extensive research and expert domain-specific knowledge \cite{dsmile}. Because of the issues mentioned above, traditional methods become slow, limits the automation process, and achieves poor generalization ability. Moreover, the overall performance of recognition broadly depends on the availability of many independent pre-processing steps. 

Here, we propose an approach that elegantly solves the problems and encapsulates all the broken-up pieces of traditional techniques into a single, unified deep neural network called `\textit{RealSmileNet}'. Our method is end-to-end trainable, fully automated, fast, and promotes real-time recognition. We employ shared convolution neural networks (CNN) layers to learn the feature extraction process automatically, Convolutional Long Short Term Memory network (ConvLSTM) \cite{ConvLSTM} layers to track the discriminative features across frames and a classification sub-network to assign a prediction score.  Our model adaptively searches, tracks, and learns both spatial and temporal facial feature representations across the frames in an end-to-end manner. In this way, the spontaneous smile's recognition becomes as simple as a forward pass of the smile video through the network. In Fig. \ref{fig:motivation}, we illustrate the difference between our method and the methods in the literature. Experimenting with four well-known smile datasets, we report state-of-the-art results without compromising the automation process. 

\noindent Our main contributions are summarized below:
\begin{itemize}
\item To the best of our knowledge, we propose the first end-to-end deep network for recognition of spontaneous and posed smiles.
\item Our method is fully automated and requires no manual landmark annotation or feature engineering. Unlike traditional methods, the proposed network can handle variable length of smile videos leading to a robust solution.
\item As a simple forward pass through the network can perform the recognition process, our approach is fast enough to promote a real-time solution.
\item We present extensive experiments on four video smile datasets and achieve state-of-the-art performance.
\end{itemize}

\section{Related Work}
\subsubsection{Dynamics of the spontaneous smile:} The smile is the most common facial expression, and usually featured by Action Unit 6 (AU6) and Action Unit 12 (AU12) in the facial action coding {system} (FACS) \cite{facs}.
{The rise of cheek and pull of lip corners is commonly associated with a smile \cite{facs}}.  In terms of temporal dynamics, the smile can be segmented into the onset, apex, and offset phases. It corresponds to the facial expression variation from neutral to smile and then return to neutral. In physiological research on facial expressions, Duchenne defines the smile as the contraction of both the zygomatic major muscle and the orbicularis oculi muscle, which known as \emph{D-Smile}. A \emph{Non-D-smile} tends to be a polite smile where only the zygomatic muscle is contracted \cite{dsmile}. Recently, Schmidt et al. \cite{DMarker} proposed a quantitative metric called Duchenne Marker (D-Marker) to measure the enjoyment of smiles. Much research uses this (controversial) metric to recognise spontaneous and posed smiles \cite{SM:2,SM:4,SM:6,DMarker}. Our end-to-end network for spontaneous smile recognition does not use the D-Maker feature.

\noindent\textbf{Spontaneous smile engineering:} The literature of spontaneous smile recognition usually follows a feature-based approach. Those methods extract features from each frame along the time dimension to construct a multi-dimensional signal. A statistical summary of the signal obtained from a smile video, such as duration, amplitude, speed, accelerations, and symmetry, is considered in smile classification. The majority of competitive and notable research on Smile Classification relies on feature extraction by D-marker \cite{SM:2,SM:4,DMarker}.  Dibeklioglu \textit{et al.} \cite{SM:2} proposed a linear SVM classifier that uses the movement signal of eyelid, lip, and cheek.  Mandal \textit{et al.} \cite{SM:4} proposed a two-stream fusion method based on the movement of eyelid and lip and the dense optical flows with SVM. Pfister \textit{et al.} \cite{CLBPTOP} proposed feature extraction by using appearance-based local spatial-temporal descriptor (CLBP-TOP) for genuine and posed expression classifications. The CLBP-TOP is an extension of LBP, which able to extract the temporal information. Later, Wu \textit{et al.} \cite{SM:5} used CLBP-TOP feature on the region of interests (eyes, lips and cheek) to train SVM for smile classifications. Valstar \textit{et al.} \cite{action} introduced the geometric feature-based smile classifications, using the movement of the shoulder and facial variation. We identify a few drawbacks of features based approaches. First, strongly dependence on accurate localization of the action units. Second, some approaches require manual labeling to track the changes in facial landmarks. Third, spontaneous smile recognition becomes a costly process - requiring laborious feature engineering and careful pre-processing.

\noindent\textbf{End-to-end solution:}  In recent decades, the advancement of graphic processing units and deep learning methods allow end-to-end learning of deep neural networks that achieves unprecedented success in object re-identification \cite{reidentification}, detection \cite{objectdetection,rahman2020any}, segmentation \cite{segmentation} using the image, videos, and 3D point cloud data \cite{pointregistration}. An end-to-end network takes the input (image/video/3D point cloud) and produces the output with a single forward pass. This network performs the feature engineering with the convolution layers and reduces the necessity of manual intervention and expert effort on the training process. In this vein, an end-to-end trainable deep learning model to automatically classify the genuine and posed smiles is the next step to solve the problems of feature-based solutions. With this motivation, Mandal \textit{et al.} \cite{accv2016} extract features from pre-trained CNN networks (VGG Face Recognition model \cite{vggface}) but eventually feed the features to a separate SVM. Instead, we propose the first fully end-to-end solution.

\section{Our Approach}

In contrast with available methods, an end-to-end deep learning model can provide a convenient solution by ensuring complete automation and saving computational cost after finishing the training. Because of the availability of enormous amounts of data in recent years, such {end-to-end} learning systems are gradually dominating research in AI. In this section, we describe an end-to-end solution for spontaneous and posed smile recognition.

\subsection{Preliminaries}
We consider a binary classification problem assigning labels to a sequence of images or video $\overrightarrow{\boldsymbol{\mathnormal {X}}_i} =  \langle  \mathbf{x}_{t} | 1 \ldots n_i \rangle$ by parameterizable models  $\mathcal{F}_{\theta}$ where, $n_i$ is number of frames associated with $\overrightarrow{\boldsymbol{\mathnormal {X}}_i}$ and $i \in \mathcal{T}$ and $\mathcal{T}$ is total number of videos in the dataset. The training dataset includes a set of tuples $\{ (\overrightarrow{\boldsymbol{\mathnormal {X}}_i}, y_i) \, : \, i \in [0,\mathcal{T}] \}$ where $y_i$ represent the ground-truth label of the $i$th video. Here,  $y_i = 1$ and  $y_i = 0$ represent the class label spontaneous / posed smile respectively. Our goal is to train an end-to-end deep network model, $\mathcal{F}_{\theta}$, that can assign a prediction label, $\hat{y}_j$, to all of $\mathcal{K}$ testing videos, $\{ \overrightarrow{\boldsymbol{\mathnormal {V}}_j} \}_{j=1}^{\mathcal{K}}$. We formulate $\hat{y}_j$ as follows:

\begin{equation}
    \hat{y}_j =
    \begin{cases}
      1, & \text{if}\ \mathcal{F}_{\theta}(\overrightarrow{\boldsymbol{\mathnormal {V}}_j}) \geq 0.5 \\
      0, & \text{otherwise}
    \end{cases}\\
    \label{eq:inference}
\end{equation}

\begin{figure}[!t]
   \centering
   \includegraphics[width=1\textwidth,trim={0cm 0cm 0cm 0cm},clip]{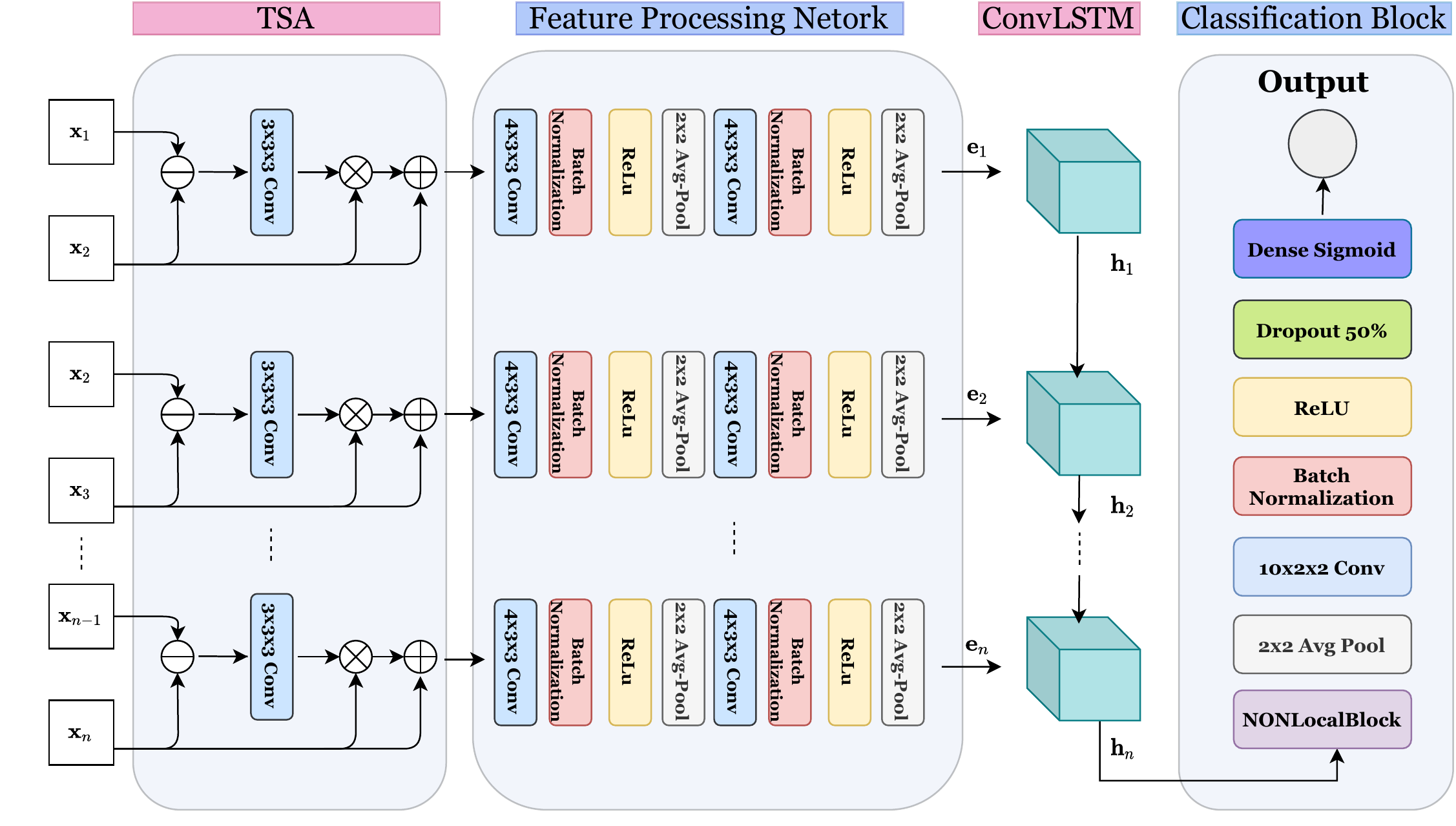}
     \caption{Our proposed \textit{RealSmileNet} architecture. The TSA layers guide the following feature processing network to extract discriminative facial representations. Then ConvLSTM tracks the face representation across the temporal direction to create a unified length video embedding. Finally, the classification block refines the video embedding and predicts a classification score. {The number of kernels is denoted by the first number of Conv block, then the size of the kernel is followed.} }
     \label{fig:arch}
\end{figure}

\subsection{Architecture}
We illustrate our proposed \textit{RealSmileNet} architecture in Fig. \ref{fig:arch}. It has four components: Temporal based Spatial Attention (TSA), Feature Processing Network (FPN), ConvLSTM, and Classification block. TSA captures the motion of frames using two consecutive frames as input, FPN further {processes} the motion feature to generate a frame representation, ConvLSTM {processes} the temporal variation of frame features across different time frame to produce a video representation, and finally a classification block predicts a label for the input video. We train all components together from scratch as a single and unified deep neural network.

\noindent\textbf{Temporal based Spatial Attention (TSA):} We design the TSA network that learns the variation of pixels i.e. motion of a video by concentrating certain regions using residual attention. Previous research on video classification \cite{Simonyan_NIPA_2014} showed that difference image of adjacent frames provides crude approximation of optical flow images that is helpful in action recognition. With this motivation, this network takes two consecutive frames of a video, applies a 2D convolution on difference map and performs some element-wise operations on the residual (skip) connections from $\mathbf{x}_t$. The overall TSA calculation is defined as:
\begin{align}
    \mathnormal{TSA(\mathbf{x}_{t-1}, \mathbf{x}_{t})} = \Big( \mathcal{C}(\mathbf{x}_{t} - \mathbf{x}_{t-1}) \otimes \mathbf{x}_{t} \Big) \oplus \mathbf{x}_{t}
\end{align}
Where, $\mathcal{C}$ represents a convolution layer that takes the difference between current frame $\mathbf{x}_{t}$ and previous frame $\mathbf{x}_{t-1}$, $\otimes$ and $\oplus$ are the {Hadamard product and element-wise addition respectively}. The residual connections augment the output of the convolution and focus on certain area in the context of the current frame.

\begin{figure}[!t]
    \centering
    \includegraphics[width= .75\linewidth]{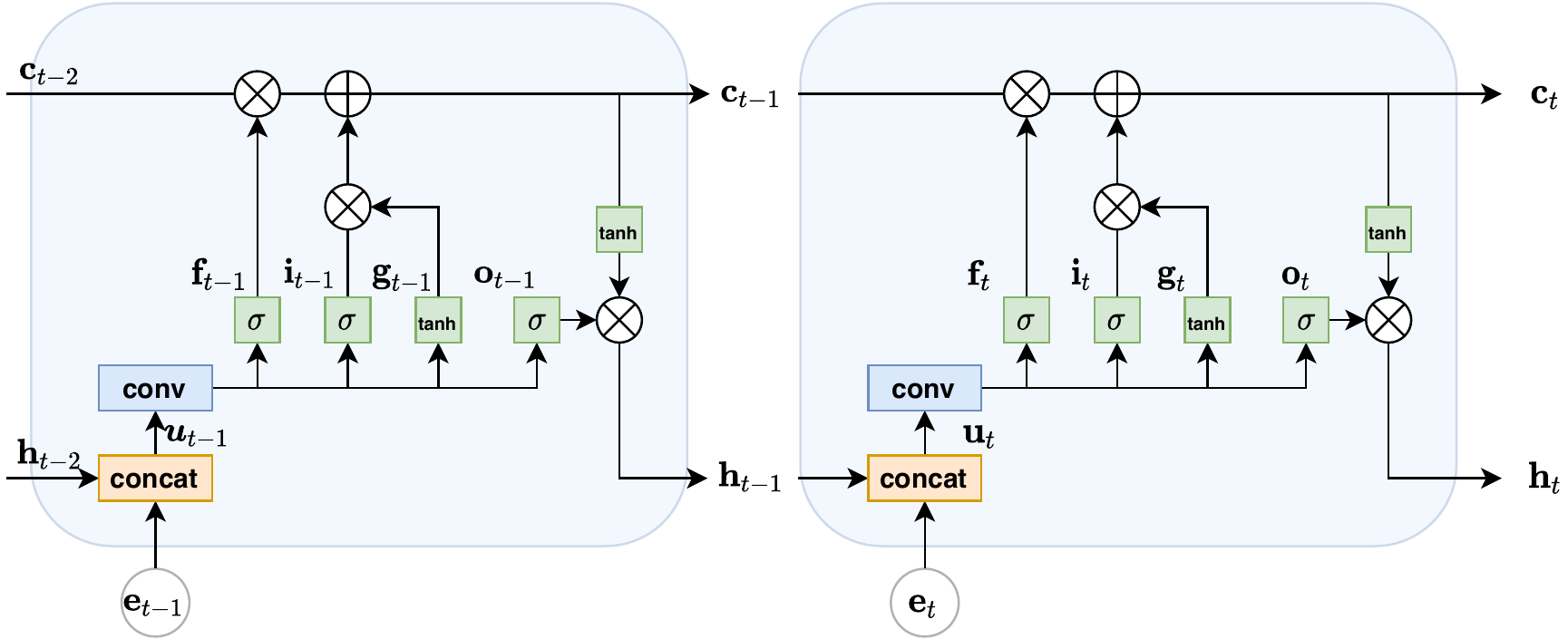}
    
    \caption{State transitions of the ConvLSTM.}
    \label{fig:ncl}
\end{figure}

\noindent\textbf{Feature Processing Network (FPN):} We forward the output of the TSA network to the FPN layers to process the TSA features further. We design FPN with two sets of Conv, Batch Normalization, ReLU, and Avg-pooling layers. In FPN block, all the convolution layer and average pooling layer use 3x3 kernel size and 2x2 kernel size respectively. FPN learns a dense spatial feature representation of frames required to model the complex interplay of smile dynamics. During the experiment, we replace this FPN with popular ResNet18 and DenseNet like structure. However, we have {achieved} the best performance using our proposed implementation of an FPN. Besides, our FPN has less trainable parameters than its alternatives. In our model, TSA and FPN contribute together to get overall spatial information from frames. This representation plays the role of D-marker \cite{SM:2,SM:4,DMarker}, HoG \cite{SM:5,CLBPTOP,accv2016}, LBP \cite{SM:5,accv2016} of the traditional approach.  The main difference is our model learns this representation, unlike conventional methods dependent on handcrafted and computationally intensive features.

\noindent\textbf{ConvLSTM:} We employ the ConvLSTM \cite{ConvLSTM} to model the temporal dynamics of the video. We adaptively build up a saliency temporal representation of each video that contributes to the classification processes. Specifically, we concurrently learn the hidden states and input tensors by using convolution layers instead of maintaining different weight matrixes for the hidden state and input. We visualize the state transition in Fig. \ref{fig:ncl} that {performs} the following operations: \textit{input vector, } $\mathbf{u}_{t} = concatenate(\mathbf{e}_{t},\mathbf{h}_{t-1})$, \textit{input gate,} $ \mathbf{i}_{t} =\sigma\left(\mathbf{W}_{i} \circledast \mathbf{u}_{t} \oplus \mathbf{b}_{i}\right)$, \textit{forget gate, } $ \mathbf{f}_{t} =\sigma\left(\mathbf{W}_{f} \circledast \mathbf{u}_{t} \oplus \mathbf{b}_{f}\right)$,  \textit{output gate, } $\mathbf{o}_{t} =\sigma\left(\mathbf{W}_{o} \circledast \mathbf{u}_{t} \oplus \mathbf{b}_{o}\right)$, \textit{cell gate, }$ \mathbf{g}_{t} =\tanh\left(\mathbf{W}_{g} \circledast \mathbf{u}_{t}  \oplus \mathbf{b}_{g}\right)$, \textit{cell state,} $ \mathbf{c}_{t} =\mathbf{f}_{t} \otimes \mathbf{c}_{t-1} \oplus  \mathbf{i}_{t} \otimes \mathbf{g}_{t} $, \textit{hidden state, } $ \mathbf{h}_{t} =\mathbf{o}_{t} \otimes \tanh \left(\mathbf{c}_{t}\right)$, where,
$\sigma$ and $\tanh$ are the activation function of the sigmoid and hyperbolic tangent. $\circledast$, $\otimes$ and $\oplus$ represent convolution operator, Hadamard product, and element-wise addition. $concatenate$ operator stands for concatenating the augments along the channel axis. $[\cdot]_{t}$ and $\mathbf{W}_{[\cdot]}$ denote the element at time slot $t$ and the corresponding weight matrix respectively. Usually, ConvLSTM updates the input gate, forget gate, cell gate, and output gate by element-wise operations \cite{ConvLSTM}. But, in our case, we learn more complex temporal characteristics by concatenating the input and hidden state as the input of the convolution layer. As nearby pixels of an image are both connected and correlated, using more complex flows within the ConvLSTM cell, the convolution layer can group the local features to generate robust temporal features while preserving more spatial information.

\noindent\textbf{Classification Block:} The hidden state of the last frame is passed to the classification block to assign a prediction label. This block is composed of dot product NonLocal block \cite{nonlocal}, average pooling (2$\times$2 kernel size), convolution layer(with 2$\times$2 kernel size), batch normalization, ReLU, Dropout (with 0.5 probability), and dense layers. The NonLocal block captures the dependency between any two positions \cite{nonlocal}. Such reliance is critical because Ekman \textit{et al.} \cite{syssmile} suggested the relative position of facial landmarks (such as symmetry) contributes to the smile classification. Then, we further trim the learned embedding of the video feature through the later layers of the classification block. In this way, the classification space is well-separated for binary classification (see Fig. \ref{fig:classificationspace}).

\noindent\textbf{Loss function:} Given a video as an input, $\overrightarrow{\mathnormal {X }}_i$, our proposed network predicts a score, $\mathcal{F}_{\theta}(\overrightarrow{\mathnormal {X }}_i)$, which is compared with the ground-truth $y_i$ to calculate the weighted binary cross-entropy loss:

\begin{equation}
    \mathcal{L}_{CE}= -\frac{1}{\mathcal{T}} \sum_{i=1}^{\mathcal{T}}\Bigg[\alpha \, y_{i} \, \log \left(\mathcal{F}_{\theta}(\overrightarrow{\mathnormal {X }}_i)\right)+ \beta \left(1-y_{i}\right) \, \log \left  (1-\mathcal{F}_{\theta}(\overrightarrow{\mathnormal { X }}_i)   \right)\Bigg],
    \label{eq:loss}
\end{equation}

where, $\alpha$ and $\beta$ are the weights computed as the proportion of spontaneous and posed videos in the training dataset respectively.

\noindent\textbf{Inference:} For $j$th test video, $\overrightarrow{\boldsymbol{\mathnormal {V}}_j}$, we perform a simple forward pass through the trained network and produce a prediction score, $\mathcal{F}_{\theta}(\overrightarrow{\boldsymbol{\mathnormal {V}}_j})$. Then, we apply Eq. \ref{eq:inference} to assign the predicted label, $\hat{y}_j$ for the input.

\begin{figure}[!tb]
    \centering
    \includegraphics[width=1\linewidth]{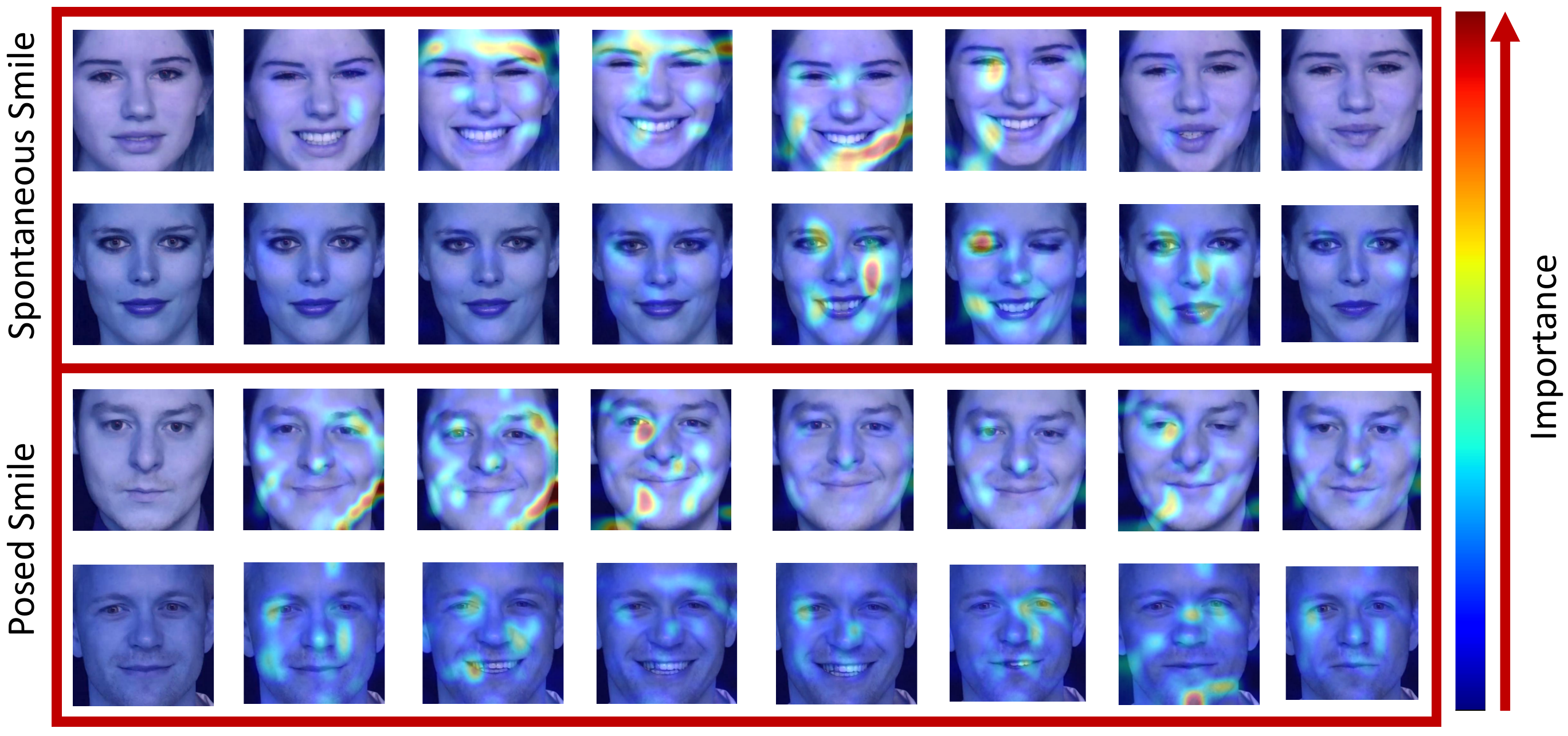}
    \caption{Visualization of FPN features for spontaneous (top two rows) and posed (bottom two rows) smiles from UVA-NEMO \cite{SM:2} by using score-cam \cite{score-cam,score-cam-vis}. Keeping equal temporal distance from each other, sample frames are selected for this visualization. The more 'warm' the color, the more important the area becomes during classification.
    } 
    \label{fig:heatmap}
\end{figure}

\subsection{Analysis}

We analyze and visualize different aspects of our model, which allows us to address many drawbacks of the traditional approaches. 

Our model automatically learns discriminative smile features that replace traditional handcrafted features. This learning process does not require manual landmark initialization and their tracking through the video. Our ConvLSTM block enables us to track the learned features automatically until the last frame. The iterative learning process of ConvLSTM does not have any restriction on the maximum length of smile videos, unlike traditional methods requiring maximum fixed video length. Our ConvLSTM block effectively manages the temporal aspect of features in the time dimension, which performs the role of face landmark tracking of other methods. Our classification block, coupled with the ConvLSTM leans to classify time series data. But, using the SVM like classifiers, which are commonly used in the area, are not an excellent fit to classify similar data. Therefore, traditional models perform hand-engineering to make the data fit for SVM. Instead, in our model, every component of traditional methods are embedded in the unified deep network. Thus, once learned, our model handles the intermediate process through a forward pass as a single unit. Such a strategy simplifies the process because that parallel implementation is easy for a deep learning model.

\begin{figure}[!t]
    \centering
   \includegraphics[width=\linewidth]{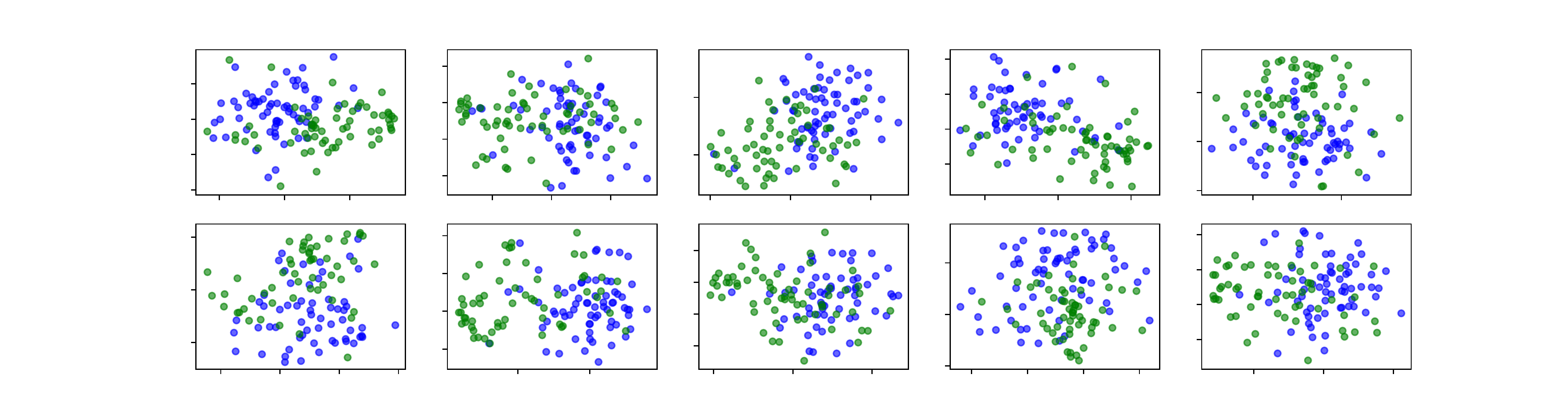}
   \caption{2D tSNE \cite{tsne} visualization of smile video features extracted from the classification block (after ReLU layer) of our proposed model. \textcolor{b} {Blue} and \textcolor{g}{Green} represent posed and spontaneous smiles respectively. Each plot shows the test fold features from the 10-fold cross-validations data of UvA-NEMO database. Here, spontaneous and posed smiles are {reasonably} well-separated to recognize using the classification block. 
   }  
\label{fig:classificationspace}
\end{figure}

\noindent\textbf{Visualization:} In Fig. \ref{fig:heatmap}, we visualize the importance of different facial regions across various frames. We extract features (after FPN layers) and blend them on the input frame. This shows that our model extracts features where it finds discriminative information. One can notice that our model puts less emphasis on neutral faces (by assigning cooler color on the heatmap) because those frames have no role in the context of spontaneous or posed smile recognition. Moreover, {the starting and ending frames (roughly, onset and offset regions) are promising to be the most discriminative for posed smiles, whereas middle frames (apex regions) are important for spontaneous smiles.} To further visualize the feature embedding, we plot learned video features (from the output of ReLU layer of classification block (please see Fig. \ref{fig:arch})) of the test fold belonging to the UvA-NEMO dataset in Fig. \ref{fig:classificationspace}. We notice that spontaneous and posed smile features are {properly} separated for classification.

\section{Experiment}

\subsection{Setup}

\textbf{Dataset:} In this paper, we experiment on four popular smile datasets: Here, we briefly describe the data statistics.  \textbf{(a) UVA-NEMO Smile Database} \cite{SM:2}\textbf{}{:} This dataset is recorded in $1920 \times 1080$ {pixels} at a rate of 50 frames per second.  It composed of 597 spontaneous and 643 posed smile videos. The length of videos distributed from 1.8 seconds to 14.2 seconds. It contains over 400 participants (185 females and 215 males) with ages from 8 years to 76 years. There are 149 young people and 251 adults. \textbf{(b) BBC database\footnote{\url{https://www.bbc.co.uk/science/humanbody/mind/surveys/smiles/}}} {\cite{SM:2,SM:5}} This dataset contains 20 videos, recorded in $314 \times 286$ {pixels} with 25 frames per second. There are 10 spontaneous and 10 posed smiles.  \textbf{(c) MMI Facial Expression Database \cite{MMI}:} They provided spontaneous and posed facial expressions separately including 38 labeled posed smiles. Apart from these posed smiles, we identified 138 spontaneous and 49 posed smile videos from 9 and 25 participants, respectively. The age of participants ranges from 19 to 64. All of the videos contain frontal recordings. The part of the spontaneous smile is in $640 \times 480$ {pixels} at 29 frames per second, and the posed smile part is recorded in $720 \times 576$ {pixels} with 25 frames per second. 
\textbf{(d) SPOS database \cite{CLBPTOP}:} It provides both gray and near-infrared sequences of images in $640 \times 480$ resolution with 25 frames per second. We use gray images in our experiments. The face region of each image has been cropped by the database owners. There are 66 spontaneous smiles, and 14 posed smiles from 7 participants. The age of participants distributed from 22 to 31, while 3 of them are male. Table \ref{databasecompare} provides a summary of these datasets.

\begin{table}[!t]
    \centering
    \scalebox{0.85}{
    \begin{tabular}{cccccccccc}
       \toprule
        \multirow{2}{*}{Database} & & \multicolumn{2}{c}{Video Spec.} & & \multicolumn{2}{c}{Number of Videos} & &
        \multicolumn{2}{c}{Number of Subjects}\\
        \cmidrule{3-4} \cmidrule{6-7} \cmidrule{9-10}
        {} && Resolution & FPS & &  Genuine  & Posed && Genuine&Posed\\
        \midrule
        UVA-NEMO && 1920 x 1080 & 50 && 597 & 643 && 357 & 368\\ \hline
        BBC && 314 x 286 & 25 && 10 & 10 && 10 & 10\\ \hline
        \multirow{2}{*}{MMI} & & 720 x 576 & 29 && \multirow{2}{*}{138} & \multirow{2}{*}{49} && \multirow{2}{*}{9} & \multirow{2}{*}{25}\\ 
        && 640 x 480 & 25 && &  &&  & \\ \hline
        SPOS && 640 x 480 & 25 && 66 & 14 && 7 & 7\\ 
        \bottomrule
    \end{tabular}}
    \caption{Summary of the smile datasets.}
    \label{databasecompare}
\end{table}

\noindent\textbf{Train/test split}: We use the standard train/test split protocol from \cite{SM:2} for UVA-NEMO database. Following the settings from \cite{SM:2,SM:5}, we perform 10-fold, 7-fold, and 9-fold cross-validation for BBC, SPOS, and MMI datasets, respectively, while maintains no subject overlap between training and testing folds. 

\noindent\textbf{Evaluation Processes:} We have evaluated our model with prediction accuracy. The accuracy is the proportion of test data that is correctly predicted by our model. We report the average result of running ten trials.


\noindent\textbf{Implementation details\footnote{Code and evaluation protocols available at: \url{https://github.com/Yan98/Deep-learning-for-genuine-and-posed-smile-classification}}:} We train our model for 60 epoch with the mini-batch size 16. To optimize network parameters, we use Adam optimizer with a learning rate $10^{-3}$ and decay 0.005. We employ weighted binary cross-entropy loss for training where the weight is the ratio between spontaneous smiles and posed smiles in training data. To prepare the video to be manageable for the network, we sample 5 frames per second, crop the face using DLIB library \cite{dlib} and resize each frame into the dimension $48 \times 48$, which are purely automatic processes. We validate the sensitivity of these design choices in experiments. We implement our method using the \textit{PyTorch} library \cite{pytorch}.

\subsection{Recognition Performance} \label{sec:results}

In this subsection, we will compare our performance with other models, will show an ablation study, will design choice sensitivity, and will analyze the robustness of our approach.

\subsubsection{Benchmark Comparisons:} 
In Table \ref{tab:overallcomparision}, we compare our performance of spontaneous and posed smile classification with previously published approaches using four popular datasets. We divide all methods into two categories: semi-automatic and fully-automatic. Semi-automatic methods manually annotate facial landmark locations of the first frame of the video. In contrast, fully-automatic methods require no manual intervention in the complete process. Our model successfully beats all methods in MMI, SPOS, and BBC datasets. For UVA-NEMO, we outperform the automatic method \cite{accv2016}. However, Wu \textit{et al.} reported the best performance on the UVA-NEMO dataset.\cite{SM:5} For all these experiments, the same video and subjects are used during testing. But, being not end-to-end, previous methods apply many pre-processing steps (borrowed from different work) that are not consistent across methods. For example, the performance of \cite{SM:2,SM:6,cohn,CLBPTOP} are adopted from the work \cite{SM:2} where same landmark initialization tracker \cite{tracker1}, face normalization, etc. are used. However, the accuracy of \cite{SM:5} and \cite{SM:4} are reported from the original papers where they employed a different manual initialization and tracker \cite{tracker3,tracker2}. Moreover, the number and position of landmarks used are also different across models. Because of these variations, performance of the semi-automatic methods are difficult to compare in a common framework. The automatic method \cite{accv2016} is our closest competitor because of the lack of requirement of landmark initialization or tracker and their best result can be gained in a fully automatic way. However, their feature extraction and learning model are still separated. Besides, to manage the variable length of video frames, they apply a frame normalization process (using fixed number of coefficients of Discrete Cosine Transform) to create fixed length videos. Our proposed model is fully-automated and end-to-end trainable as a single deep learning model. It requires no landmark detection, tracking, frame normalization to a fixed-length, etc.

\begin{table}[!t]
\renewcommand{\tabcolsep}{7pt}
    \centering
    \scalebox{0.85}{
\begin{tabular}{lccccc}
    \toprule
    \cmidrule{3-6}
     {Method} & Process Type& {}   UVA-NEMO & MMI & SPOS & BBC\\ 
    \midrule
    Cohn'04 \cite{cohn}&Semi-automatic& 77.3 & 81.0 & 73.0 & 75.0\\
    Dibeklioglu'10 \cite{SM:6} &Semi-automatic& 71.1 & 74.0 & 68.0 & 85.0 \\ 
    Pfister'11\cite{CLBPTOP} & Semi-automatic& 73.1 & 81.0 & 67.5 & 70.0\\ 
    Wu'14 \cite{SM:5} &Semi-automatic& \textbf{91.4} & - & 79.5 & \textbf{90.0}\\
    Dibeklioglu'15 \cite{SM:2} &Semi-automatic& 89.8 & 88.1 & 77.5 & \textbf{90.0}\\ 
    Mandal'17 \cite{SM:4} &Semi-automatic& 80.4 & - & - & -\\\hline
    Mandal'16 \cite{accv2016}  &Fully-Automatic& 78.1  & - & - & -\\
    Ours &Fully-Automatic& 82.1 & \textbf{92.0} & \textbf{86.2} &\textbf{90.0}\\
    \bottomrule
\end{tabular}}
    \caption{Benchmark comparison of methods. `-' means unavailable result.}
    \label{tab:overallcomparision}
\end{table}

\noindent\textbf{Design Choice Sensitivity:} In Fig. \ref{fig:sensitivity}, we report the sensitivity of the design choice of our method for different numbers of frames per second (FPS) and resolutions of the input frames. For all combination of FPS (1, 3, 5 and 7) and resolution ($48\times48, 64 \times64, 96\times96$ and $112\times112$) choices, we find FPS = 5 and resolution = $48\times48 $ achieves the maximum performance. Note that image resolution is important because it decides the type of visual features extracted by the CNN layers. For example, a low resolution ($48\times48$) lets the CNN {kernels} (of size $3\times3$)  
{extracts} coarse features that are the most discriminative for smile classification. Similarly, the choice of FPS also interacts with ConvLSTM layers to track the change of smile features across frames. The FPS = 5 and resolution = $48\times48$ is the trade-off to maximize the performance.

\begin{table}[!t]
    \begin{minipage}[!t]{0.5\linewidth}
	\scalebox{0.85}{
\begin{tabular}{lcccc}
    \toprule
    \cmidrule{2-5}
     {Method} &  {}   UVA-NEMO & MMI & SPOS & BBC\\ 
    \midrule
    No TSA & 78.5 &  92.0 & 81.5 &  80.0 \\ 
    miniResNet  & 73.8&  84.9& 80.5 &  \textbf{90.0} \\ 
    miniDenseNet   & 77.0 &  71.2 &  82.2&  \textbf{90.0} \\
    No Weighted Loss  & 80.6  & 91.7 & 82.2   &   \textbf{90.0} \\
    Softmax Function  & 79.2 & 71.6 & 82.2  &  70.0 \\ \hline
    Ours & \textbf{82.1} & \textbf{92.0} & \textbf{86.2} &\textbf{90.0} \\ 
    \bottomrule
\end{tabular}}
        \caption{Ablation study. We experiment adding or removing parts of proposed method with reasonable alternatives.}
        \label{tab:ablation}
	\end{minipage}
	\hfill 
	\begin{minipage}[!t]{0.48\linewidth}
		\includegraphics[width=1\textwidth,trim={.8cm 0cm 1cm .5cm},clip]{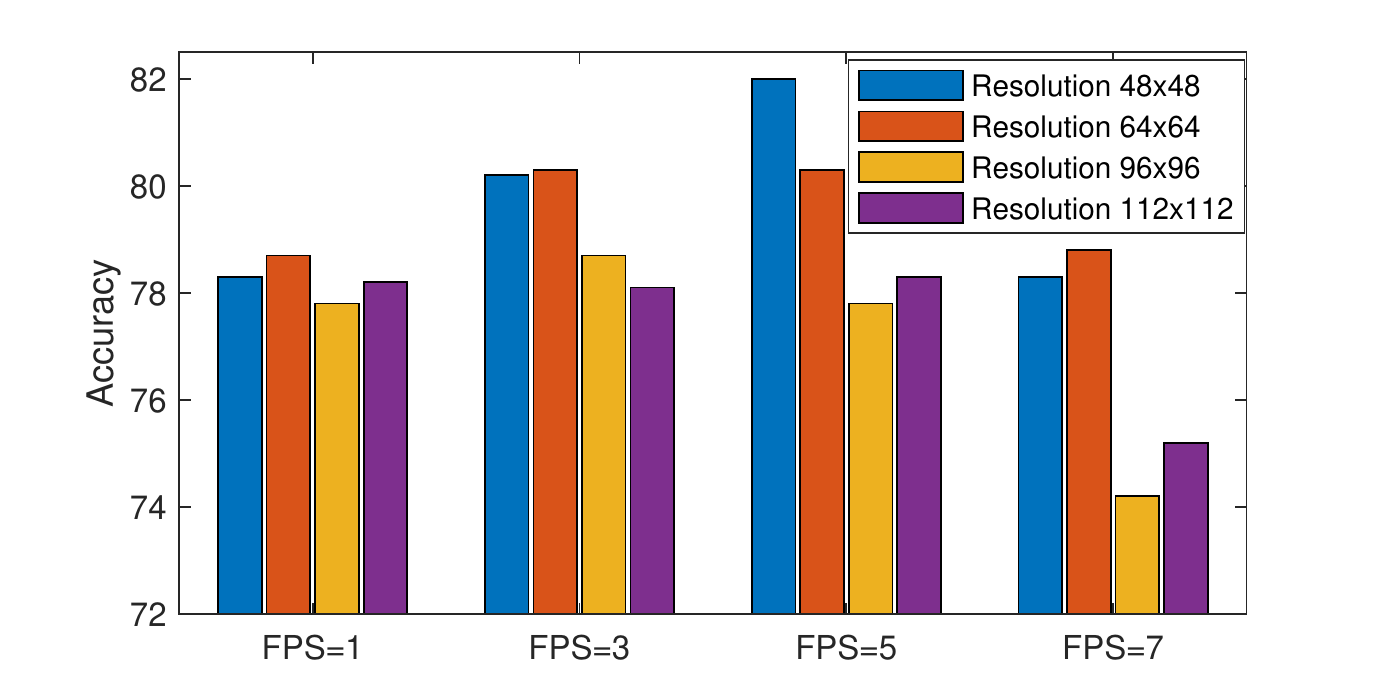}
		\captionof{figure}{Varying image size and FPS on UVA-NEMO dataset.}
		\label{fig:sensitivity}
	\end{minipage}
\end{table}

\noindent\textbf{Ablation studies:} In {Tab. \ref{tab:ablation}}, we perform ablation studies by replacing part of our proposed network with a suitable alternative. Our observations from the ablation studies are the following. \textbf{(\textit{1})} We remove the TSA block from our model and directly forward the frame to FPN for feature extraction. In this situation, the network could not get optical flow information and motion from the spatio-temporal region. Smile features based on the difference of consecutive frames extract more discriminative features than a single frame. Thus, without using the TSA block, the performance degrades, especially on UVA-NEMO and SPOS datasets. \textbf{(\textit{2})} Now, we experiment on the alternative implementation of the FPN network, for example, ResNet12 \cite{resnet} (composed by 3 layers) or DenseNet \cite{densenet} (with growth rate 2 and 6,12, 24 and 16 blocks in each layer). We use a relatively small version of that well-known architecture because smile datasets do not contain enough instances (in comparison to large scale ImageNet dataset \cite{ILSVRC_2015}) to train large networks. Alternative FPNs could not outperform our proposed FPN implementation. One reason could be that the smaller version of those popular architectures {is} still larger than our proposed FPN, and available smile data {overfits} the networks. Another reason is that, we could not use pre-trained weights for the alternatives, because of different input resolutions. \textbf{(\textit{3})} We try without the weighting version of the loss of Eq. \ref{eq:loss}, i.e, $\alpha = \beta = 1$. This impacts the performance of UVA-NEMO, MMI, and SPOS dataset because of the large imbalance in number of training samples of spontaneous and posed smiles. \textbf{(\textit{4})} We replace the dense sigmoid at the last layer of the classification block with softmax. In our sigmoid based implementation, we use one neuron at the last layer to predict a score within $[0,1]$ and apply Eq. \ref{eq:inference} for inference. For the softmax case, we add two neurons at the last layer, which increases the number of trainable parameters. We notice that the softmax based network does not perform better than our proposed sigmoid based case. This observation is in line with the recommendation of \cite{actfun} that Softmax is better for the multi-class problem rather than a binary class case. \textbf{(\textit{5})} The performance of our final model outperforms all ablation alternatives consistently across datasets. 

\begin{figure}[!t]
  \centering
\includegraphics[width=.8\textwidth,trim={1.75cm 0cm 4cm .2cm},clip]{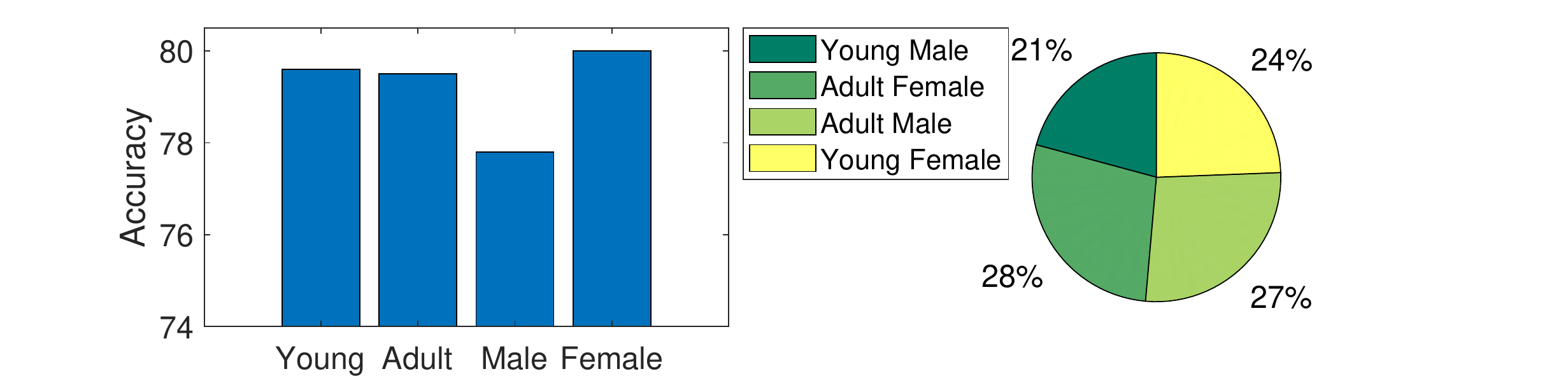}
\caption{\textbf{\textit{(left)}} The accuracy of the model trained by individual group. \textbf{\textit{(right)}} The normalized distribution of the total number of wrong predictions among different subject groups.
}
\label{fig:ageandgender}
\end{figure}

\noindent\textbf{Effect of Age and Genders:} To illustrate our approach's robustness, we analyze the effect of model prediction for the different subject groups concerning age and gender.  Firstly, we experiment on whether biologically similar appearance inserts any bias in the prediction.  For this, we train and test our proposed model with only male/female/adult/young separately. In Fig. \ref{fig:ageandgender}\textbf{(\textit{left})}, we show the results on each subgroup: male, female, adult, and young. We notice that the performance is similar to our overall performance reported in Table \ref{tab:overallcomparision}. This shows that our training has no bias on age- and gender-based subgroups. Secondly, in Fig. \ref{fig:ageandgender}\textbf{\textit{(right)}}, we visualize the normalized distribution of the total number of wrong predictions among adults/young and males/females. We find that there is no significant bias in misprediction distribution. In other words, as the mispredictions are similar across different groups, our model does not favor any particular group.

\noindent\textbf{Cross-domain Experiments:}{ We also perform experiments across datasets and subject groups. While training with UVA-NEMO and testing with BBC, MMI, and SPOS dataset, we get 80\%, 92.5\% \& 82.5\% accuracy, respectively. Moreover, training with adults and testing with young subjects got 75.9\%, and conversely 74.8\% accuracy. Again, training on female then testing on male subjects got 74.2\% and conversely 75.1\% accuracy. These experiments indicate the robustness of our method.}

\subsection{Discussion}

\noindent\textbf{Time Complexity:} The processes of facial landmarks tracking/detection followed by handcrafted feature extraction are usually very computationally expensive. As evidence, when we re-implement D-marker feature-based approaches with DLIB library \cite{dlib} to face normalization and facial landmark detection, it requires more than 28 hours for the processes using a single NVIDIA V100 GPU and one Intel Xeon Cascade Lake Platinum 8268 (2.90GHz) CPU. Although the training is efficient and effective, the pre-processing pipelines are costly. However, for our end-to-end learning models, the whole processing only spends up to eight hours using the same system configuration, which significantly saves time.

\begin{wrapfigure}{R}{0.5\textwidth}
  \begin{center}
    \includegraphics[width=.5\textwidth,trim={0cm 0cm .8cm 0cm},clip]{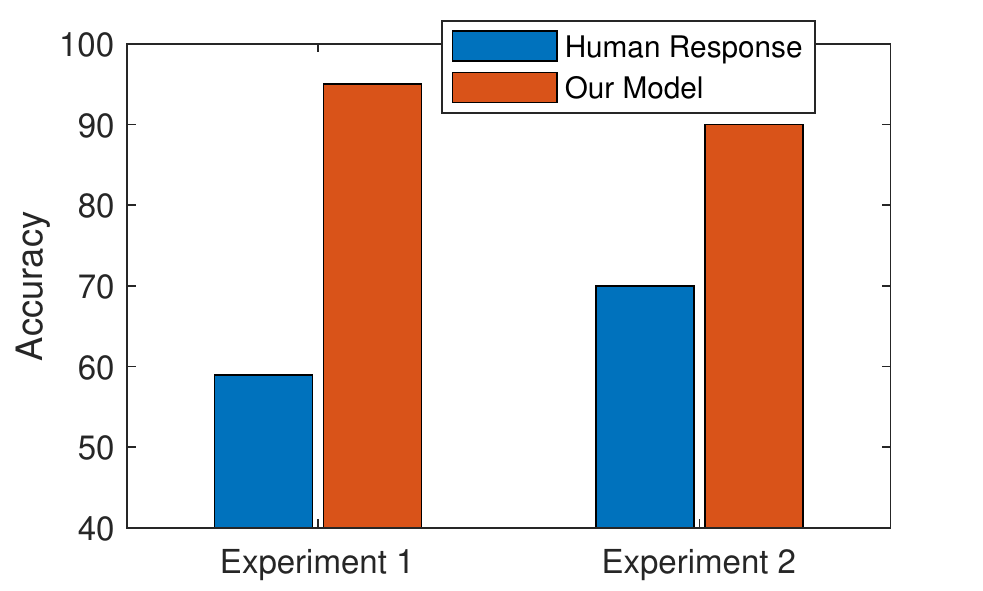}
    \end{center}
    \caption{Human response vs. our model}
    \label{fig:human}
\end{wrapfigure}

\noindent\textbf{Human Response vs. Our Model:} Several recent works estimate the ability of the average human to recognize spontaneous smiles \cite{zk3,zk4}. In Fig. \ref{fig:human}, we show the comparison of our work with human responses using the experiment set-up mentioned in \cite{zk3}. In an experiment with 26 human and 20 videos selected from the UVA-NEMO dataset, Hossain \textit{et al.} \cite{zk3} reported a $59\%$ average accuracy of the human. In this set-up, our trained \textit{RealSmileNet} (without using any of those 20 videos and their subjects during training) successfully achieves $95\%$ of accuracy on the same test set. In another experiment with 36 humans and 30 videos from the same UVA-NEMO dataset, Hossain \textit{et al.} \cite{zk3} reported $70\%$ for humans, whereas our proposed model achieves  $90\%$ accuracy. These comparisons show that our end-to-end model is already capable of beating human-level performance.

\noindent\textbf{Limitations:} One notable drawback of deep learning-based solutions is the dependence on large-scale balanced data. Being a deep model, our proposed model has also experienced this issue during training with UVA-NEMO dataset, which includes smiles of subjects with a wide range of ages, e.g., child (0--10 years), young (11--18 years), adult (19--69 years), aged people ($\geq$ 70 years).  However, among the 1,240 videos, the distributions are 20.24$\%$, 18.47$\%$, 60.16$\%$, and 1.45$\%$ respectively. The imbalanced/skewed distribution usually cannot be well modeled in the deep models \cite{skd}, and can lead to unexpected bias in the kernel of the convolution layer. Here, our model performs less well than the semi-automatic method of Wu \textit{et al.} \cite{SM:5} (See Table \ref{tab:overallcomparision}). In future, one can collect more data to handle such shortcomings.

\section{Conclusion}
Traditional approaches for recognizing spontaneous and posed smiles depend on expert feature engineering, manual labeling, and numerous costly pre-processing steps. In this paper, we introduce a deep learning model, \textit{RealSmileNet}, to unify the broken-up pieces of intermediate steps into a single, end-to-end model. Given a smile video as input, our model can generate an output prediction by a simple forward pass through the network. Our proposed model is not only fast but also removes the expert intervention (hand engineering) in the learning process. Experimenting with four large scale smile datasets, we establish state-of-the-art performances on three datasets. Our experiment previews the applicability of \textit{RealSmileNet} to many real-word applications like polygraphy, human-robot interactions, investigation assistance, and so on.

\noindent\textbf{Acknowledgment:} This research was supported by the National Computational Infrastructure (NCI), Canberra, Australia.

\bibliographystyle{splncs}
\bibliography{accv2020cameraready}

\end{document}